\documentclass[11pt]{amsart}
\usepackage{fullpage}
\usepackage[T1]{fontenc}
\usepackage[utf8]{inputenc}
\usepackage{lmodern}
\usepackage{graphicx}
\usepackage{amsmath,amssymb,amsfonts}
\usepackage{booktabs}
\usepackage{makecell}
\usepackage{microtype}
\usepackage{xcolor}
\usepackage{float}
\usepackage{subcaption}
\usepackage[section]{placeins}
\usepackage{tikz}
\usetikzlibrary{arrows.meta,positioning}
\usepackage[hidelinks]{hyperref}

\title[Structural Difficulty in Arithmetic Puzzles]{4OPS: Structural Difficulty Modeling in Integer Arithmetic Puzzles}

\author[Y. E. Zeytuncu]{
Yunus E. Zeytuncu}

\address{University of Michigan-Dearborn, Dearborn, MI, USA}
\email{zeytuncu@umich.edu}
\date{Preprint. Accepted at AIED 2026.}

\begin{document}

\maketitle

\begin{abstract}
Arithmetic puzzle games provide a controlled setting for studying difficulty in mathematical reasoning tasks, a core challenge in adaptive learning systems. We investigate the structural determinants of difficulty in a class of integer arithmetic puzzles inspired by number games. We formalize the problem and develop an exact dynamic-programming solver that enumerates reachable targets, extracts minimal-operation witnesses, and enables large-scale labeling.

Using this solver, we construct a dataset of over 3.4 million instances and define difficulty via the minimum number of operations required to reach a target. We analyze the relationship between difficulty and solver-derived features. While baseline machine learning models based on bag- and target-level statistics can partially predict solvability, they fail to reliably distinguish easy instances. In contrast, we show that difficulty is fully determined by a small set of interpretable structural attributes derived from exact witnesses. In particular, the number of input values used in a minimal construction serves as a minimal sufficient statistic for difficulty under this labeling.

These results provide a transparent, computationally grounded account of puzzle difficulty that bridges symbolic reasoning and data-driven modeling. The framework supports explainable difficulty estimation and principled task sequencing, with direct implications for adaptive arithmetic learning and intelligent practice systems.

\end{abstract}

\section{Introduction}\label{sec:introduction}

Adaptive learning systems depend critically on the ability to select, sequence, and explain practice tasks in ways that support learning progress. In domains such as arithmetic and early algebra, learners’ trajectories are shaped not only by feedback correctness but by whether tasks are appropriately challenging: tasks that are too easy provide limited learning benefit, while tasks that are too difficult can lead to disengagement or unproductive guessing. Despite extensive work on knowledge tracing and item-response models, task difficulty is often treated as an empirical label inferred from performance data, offering limited insight into \emph{why} a task is difficult or how difficulty relates to the structure of the underlying reasoning process.

In this paper, we investigate an alternative perspective: in reasoning-intensive domains, task difficulty can be defined as a \emph{structural property} of the solution space. Specifically, we show that for a class of integer arithmetic puzzles, difficulty is determined by a minimal structural invariant derived from exact solution witnesses. Using an exact symbolic solver, we enumerate valid constructions and extract \emph{minimal solution witnesses}, which reveal the compositional structure required to reach a target.

Our central finding is that the minimal number of inputs required in any valid solution, which we term \emph{minimal input usage}, fully determines difficulty under the proposed labeling scheme. Empirically, we show that this quantity acts as a minimal sufficient representation of difficulty: once minimal input usage is known, additional surface features provide no further predictive power for difficulty classification. This result establishes a direct link between symbolic structure and difficulty, providing a computationally grounded and interpretable account of task complexity.

We develop this framework in the context of arithmetic reasoning tasks inspired by number games, where each instance consists of a multiset of integers and a target value, and solutions are valid expressions under integer constraints. Using our solver, we construct a dataset of over 3.4 million instances with exact difficulty labels. We then analyze how structural properties of solution spaces relate to difficulty and evaluate the extent to which these properties can be recovered by learning-based models.

Our goal is not to replace performance-driven models of learner knowledge, but to complement them with solver-grounded semantics that support (i) principled task selection and sequencing, (ii) interpretable explanations of difficulty, and (iii) data-efficient modeling that respects symbolic constraints. While our current study focuses on structural difficulty defined by solution properties, an important direction for future work is to evaluate how these measures align with human-perceived difficulty in real learning environments.

\paragraph{Contributions.}
This paper makes four main contributions:
(1) An exact symbolic framework for enumerating arithmetic puzzle solution spaces and extracting minimal witnesses.
(2) A dataset of over 3.4 million instances with solver-grounded labels for solvability and difficulty.
(3) A structural characterization of difficulty, showing that minimal input usage is a minimal sufficient statistic.
(4) An interpretable and deployable approach to difficulty estimation that supports adaptive sequencing in learning systems.

\section{Educational Setting and Research Questions}

We consider arithmetic reasoning tasks as practice items within an adaptive learning setting, where learners construct expressions to reach a target value using a given set of integers. In such settings, adaptive systems must decide which tasks to present next, how to scaffold problem solving, and how to justify progression in ways that are meaningful to learners and instructors.

This motivates the following research questions:

\textbf{RQ1 (Structural definition).} Can task difficulty for arithmetic reasoning problems be defined in a way that is computationally exact and interpretable, based on the structure of valid solutions rather than observed performance?

\textbf{RQ2 (Determinants of difficulty).} Which solver-derived structural properties of solution spaces explain differences in difficulty, and to what extent can difficulty be reduced to a minimal structural representation?

\textbf{RQ3 (Implications for adaptivity).} How can solver-grounded difficulty measures support adaptive sequencing and task selection, enabling principled progression and interpretable recommendations?

\section{Related Work}
\label{sec:related-work}

\paragraph{Difficulty modeling in adaptive learning.}
Modeling and controlling task difficulty is central to personalization in adaptive educational systems. Early work in adaptive hypermedia focused on selecting and organizing content based on inferred learner needs \cite{brusilovsky2001adaptive}, while educational measurement frameworks treat item difficulty as a key parameter for interpreting performance \cite{embretson2000item}. More recent systems infer difficulty from response data such as accuracy or latency, which can obscure the structural sources of task complexity. Our work complements these approaches by defining difficulty directly from solution structure, independent of observed learner behavior.

\paragraph{Interpretability in educational systems.}
Interpretability is critical in educational settings, where system decisions must be understandable to instructors and learners. While post hoc explanation methods are common, there is increasing emphasis on inherently interpretable models \cite{rudin2019stop}. By grounding difficulty in solver-derived structural attributes, our approach provides transparent explanations of why certain tasks are more demanding in terms of compositional reasoning and coordination.

\paragraph{Hybrid symbolic--statistical approaches.}
Hybrid systems that combine symbolic reasoning with statistical learning aim to achieve both structural guarantees and empirical adaptability \cite{garcez2019neural}. In educational applications, symbolic components can capture domain constraints and solution structure, while statistical models account for variability in learner behavior. Our framework follows this paradigm by using an exact solver to generate verifiable difficulty semantics, which can then be approximated or incorporated into learning-based models for adaptive decision-making.

\section{Exact Solver and Search-Space Characterization}

We formalize the arithmetic puzzle setting and develop an exact symbolic solver that characterizes reachable values and minimal constructions. The solver provides ground-truth labels for solvability and difficulty, and exposes structural properties used in subsequent analysis.

\subsection{Problem Definition}

A 4OPS instance consists of a multiset of positive integers
\[
B = \{a_1, a_2, \dots, a_n\}
\]
and a target integer $T$. The goal is to determine whether $T$ can be obtained using $+, -, \times, /$ under integer-only constraints: each number is used at most once, all intermediate values are positive integers, subtraction results remain positive, and division must be exact.

We adopt a subset-allowed formulation, where solutions are not required to use all elements of $B$, enabling finer structural distinctions.

\subsection{State Representation and Closure}

States are represented as sorted tuples of integers to allow canonical memoization. From a state $S = (x_1, \dots, x_k)$, all unordered pairs $(x_i, x_j)$ are combined using valid operations to generate successor states.

For each input $B$, we compute the closure of reachable values via memoized search, storing
\[
R(S): v \mapsto \text{minimum number of operations required to obtain } v.
\]

\subsection{Witness-Based Dynamic Programming}

To capture solution structure, we reconstruct minimal expressions using subset-indexed dynamic programming. For each subset $S \subseteq \{1, \dots, n\}$, we compute
\[
DP[S](v) = (\text{min ops}, \text{expression}),
\]
representing the optimal way to obtain $v$ using exactly the elements in $S$.

Each subset is built from disjoint partitions $S = A \cup B$, combining values from $DP[A]$ and $DP[B]$. This formulation yields minimal witnesses and identifies which inputs are used in optimal constructions.

\subsection{Structural Signals}

The solver outputs reachable values, minimum operation counts, and minimal witnesses. From these, we extract structural signals, most notably the number of inputs used in a minimal solution. This quantity captures the degree of compositional coordination required and serves as the primary determinant of difficulty in later analysis.

\section{Dataset Construction}

We construct a large-scale dataset of arithmetic puzzle instances with solver-derived labels, emphasizing coverage and reproducibility.

\subsection{Bag Enumeration}

We consider six-number multisets consisting of five single-digit integers (from $\{1, \dots, 9\}$, with repetition allowed) and one value from $\{25, 50, 75\}$. Bags are represented as sorted tuples to remove permutations, yielding 3,861 distinct inputs.

\subsection{Target Space and Labeling}

For each bag, we evaluate all three-digit targets $T \in \{100, \dots, 999\}$. Using the exact solver, we compute:
\begin{itemize}
\item solvability (reachable vs.\ unreachable),
\item minimum operation count for reachable targets,
\item structural attributes derived from minimal witnesses.
\end{itemize}

This results in over 3.4 million labeled instances grounded in exact symbolic computation.

\subsection{Preliminary Difficulty Labels}

We define coarse difficulty categories based on minimum operation count:
\begin{itemize}
\item Unsolvable: target not reachable,
\item Easy: 0--2 operations,
\item Medium: 3--4 operations,
\item Hard: 5 operations.
\end{itemize}

While these labels provide a well-defined baseline, they do not capture structural differences in solution construction. In particular, operation count alone does not reflect how many inputs must be coordinated. In the next sections, we show that incorporating subset-level structure yields a more precise and interpretable characterization of difficulty.

\section{Descriptive Analysis and Baseline Modeling}\label{sec:analysis}

We analyze the dataset to characterize solvability and difficulty distributions, and to evaluate how well solver-independent features capture these properties.

The dataset contains 3,474,900 (bag, target) instances generated from 3,861 distinct bags and all three-digit targets. Under subset-allowed rules, approximately 87\% of instances are solvable, with substantial variability across bags, indicating that input composition plays a key role in determining reachability.

Difficulty labels based on minimum operation count are highly imbalanced. Easy instances (0--2 operations) are relatively rare, while medium (3--4 operations) and hard (5 operations) dominate. This reflects the combinatorial nature of the search space, where short constructions are uncommon.

We evaluate two baseline tasks: solvability prediction and difficulty classification, using interpretable models trained on solver-independent features derived from the input bag and target. Logistic regression achieves approximately 90\% accuracy on solvability prediction, with high recall for reachable targets, indicating that coarse features capture much of the reachability structure.

In contrast, difficulty prediction is substantially harder. A gradient-boosted classifier achieves approximately 73\% accuracy, performing well on medium and partially on hard instances, but consistently failing to identify easy cases. This failure suggests that easy instances are not characterized by surface-level statistics, but instead depend on finer structural properties of the solution space.

These results indicate that minimum operation count alone provides an incomplete account of difficulty. In particular, distinguishing easy instances requires information about how solutions are constructed. This motivates the use of solver-derived structural features, which we show in the next section fully determine difficulty.

\section{Solver-Derived Structural Features}
\label{sec:structural}

The baseline results in Section~\ref{sec:analysis} show that solver-independent features fail to reliably identify easy instances, indicating that difficulty depends on finer structural properties of solutions. To address this, we extract structural features from minimal solution witnesses computed by the exact solver.

For each solvable instance, we consider features derived from minimal constructions, including the number of inputs used (subset size), operator usage, intermediate magnitudes, and the number of distinct minimal solutions. These features capture aspects of compositional structure that are not visible from surface-level statistics.

\subsection{Structural Determination of Difficulty}

When structural features are incorporated into the difficulty prediction task, classification performance improves dramatically. In particular, models augmented with solver-derived features achieve perfect accuracy on the solver-defined difficulty labels.

This result reflects a deeper property of the problem: difficulty, as defined by minimum operation count, is fully determined by structural attributes of minimal witnesses. Among these, a single feature dominates.

\subsection{Minimal Input Usage as a Sufficient Statistic}

Ablation analysis reveals that the number of inputs used in a minimal witness (subset size) alone is sufficient to recover the difficulty labels exactly. Adding this feature to the baseline model yields perfect classification accuracy across all difficulty classes, and no additional features improve performance further.

This establishes subset size as a \emph{minimal sufficient statistic} for difficulty under the proposed labeling scheme. Structurally, easy instances admit solutions using small subsets of inputs, while hard instances require combining most or all available values.

\subsection{Interpretation and Educational Significance}

Subset size provides a direct and interpretable measure of compositional demand. From a learner perspective, it reflects how many quantities must be simultaneously coordinated during problem solving. Instances solvable with fewer inputs often admit direct strategies, while those requiring more inputs impose greater working-memory and integration demands.

This connection provides a principled bridge between solver-derived structure and cognitive notions of difficulty, explaining why surface-level features fail to capture easy cases and why structural attributes are necessary.

\subsection{Implications for Adaptive Systems}

Because difficulty is structurally determined, adaptive systems can use minimal input usage as a transparent control variable for sequencing. Tasks can be ordered by increasing subset size, yielding progressions that are both principled and explainable. This allows systems to distinguish between task complexity and learner proficiency, enabling more targeted support.

\subsection{Scope and Limitations}

The results hold within the current formulation of integer arithmetic puzzles and solver-defined difficulty. While subset size captures structural necessity, it does not account for all factors influencing human performance, such as fluency, strategy familiarity, or affective variables. Evaluating alignment between structural difficulty and human-perceived difficulty remains an important direction for future work.

\section{Educational Implications and Discussion}
\label{sec:educational-implications}

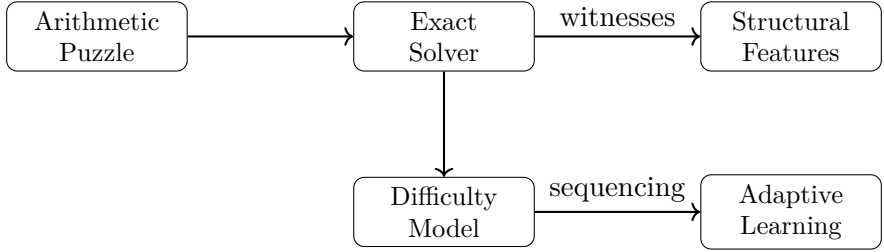
\begin{figure}[t]
\centering
\begin{tikzpicture}[
    box/.style={draw, rectangle, rounded corners, align=center,
                minimum width=2.4cm, minimum height=0.9cm, font=\small},
    arrow/.style={->, thick},
    node distance=1.4cm and 2.2cm
]

\node[box] (puzzle) {Arithmetic\\Puzzle};
\node[box, right=of puzzle] (solver) {Exact\\Solver};
\node[box, right=of solver] (structure) {Structural\\Features};

\node[box, below=of solver] (difficulty) {Difficulty\\Model};
\node[box, right=of difficulty] (adaptive) {Adaptive\\Learning};

\draw[arrow] (puzzle) -- (solver);
\draw[arrow] (solver) -- node[above]{witnesses} (structure);
\draw[arrow] (solver) -- (difficulty);
\draw[arrow] (difficulty) -- node[above]{sequencing} (adaptive);

\end{tikzpicture}
\caption{Overview of the proposed framework. Exact symbolic solving yields interpretable structural features that explain puzzle difficulty and support adaptive learning applications.}
\label{fig:pipeline}
\end{figure}

This work provides a principled foundation for difficulty modeling in arithmetic learning systems by grounding difficulty in solver-derived structural properties. Figure~\ref{fig:pipeline} summarizes the proposed framework: exact symbolic solving produces minimal witnesses and structural descriptors, which explain difficulty and support adaptive sequencing.

Unlike black-box difficulty predictors, this approach is inherently interpretable. Difficulty is explained in terms of concrete algebraic structure, allowing instructors and system designers to understand \emph{why} a problem is easy or hard.

A central finding is that difficulty is governed primarily by the number of inputs required in a minimal construction. Problems solvable using small subsets of inputs tend to be easier, while those requiring most or all inputs exhibit greater compositional complexity. This yields a natural and transparent ordering of tasks, enabling adaptive systems to sequence problems based on structural demand rather than surface features.

From a systems perspective, the separation between solver-based analysis and lightweight prediction provides a practical deployment pathway. Exact solvers can be used offline to generate labeled datasets and extract structural regularities, while learned models approximate difficulty online for real-time adaptation. This hybrid symbolic--statistical approach aligns with the broader goal of building adaptive systems that are both effective and interpretable.

Future work will examine how solver-derived difficulty aligns with human performance and perceived challenge. The 4OPS puzzle environment enables the collection of large-scale gameplay data to study behavioral difficulty, strategy formation, and learner adaptation.

More broadly, this work demonstrates that difficulty in structured problem-solving domains can be characterized through minimal solution structure. Rather than relying on surface complexity, difficulty emerges from structural necessity. This perspective offers a pathway toward interpretable and principled difficulty modeling in adaptive learning systems.

\section{Conclusion}

We investigated the problem of modeling task difficulty in arithmetic reasoning environments through a structural lens. Using an exact symbolic solver, we constructed a dataset of over 3.4 million puzzle instances and extracted minimal solution witnesses that reveal the compositional structure of valid constructions.

Our central result is that difficulty, defined by minimum operation count, is fully determined by a simple structural invariant: the number of inputs required in a minimal solution. This quantity is a minimal sufficient statistic for difficulty, showing that task complexity arises from structural necessity rather than surface-level properties.

This perspective provides a transparent and computationally grounded foundation for difficulty estimation. In contrast to purely data-driven approaches, it enables interpretable explanations and supports principled task sequencing based on structural demand. The underlying puzzle game has been deployed as a free mobile application, providing an accessible platform for studying arithmetic reasoning and adaptive difficulty in real-world settings.

More broadly, this work demonstrates how exact symbolic reasoning can inform learning-relevant constructs in adaptive systems. Future work will focus on validating the alignment between structural difficulty and human performance, and on extending these ideas to richer domains where solution structure governs problem complexity.

\section*{Acknowledgments}

The author thanks Omer Aydas for his contributions to the development of the underlying arithmetic puzzle game that motivated this work.

\end{document}